\documentclass[11pt]{article}

\usepackage[final]{acl}

\usepackage{times}
\usepackage{latexsym}

\usepackage[T1]{fontenc}

\usepackage[utf8]{inputenc}

\usepackage{microtype}
\usepackage{booktabs}
\usepackage{amsmath}

\usepackage{subcaption}
\usepackage{booktabs}
\usepackage{graphicx}
\usepackage[table]{xcolor} 

\definecolor{lgray}{gray}{0.9} 

\usepackage{multirow}
\usepackage{booktabs}

\usepackage{inconsolata}

\usepackage{graphicx}

\title{What You Post Is What You Symptom: Estimating Psychometric Networks from Social Media Discourse}

\author{
  {\bf Ștefana-Arina Tăbușcă}$^{1}$ \quad {\bf Ana Sabina Uban}$^{2,3}$ \quad {\bf Liviu P. Dinu}$^{2,3}$ \\
  \small $^{1}$Interdisciplinary School of Doctoral Studies \quad $^{2}$Faculty of Mathematics and Computer Science \\
  \small $^{3}$Human Language Technologies Research Center, University of Bucharest, Romania \\
  \small \texttt{stefana.tabusca@s.unibuc.ro} \quad \texttt{auban@fmi.unibuc.ro}  \quad \texttt{ ldinu@fmi.unibuc.ro}
}

\begin{document}
\maketitle
\begin{abstract}
Symptom networks in psychopathology are estimated almost exclusively from self-report questionnaires, restricting inference to whatever symptom set a given instrument covers. We introduce a framework for symptom-resolved network estimation directly from social media discourse and apply the framework to depression and eating disorders, with a third anorexia corpus as cross-dataset validation. To our knowledge, we are the first to estimate symptom-resolved psychometric networks from social media text, compare them to clinical co-endorsement networks, and replicate transdiagnostic linguistic markers across 3 independently recruited corpora spanning 2 disorders simultaneously. For depression, expression scores correlate with BDI-II totals at $\rho{=}0.434$, 20 of 21 items reach FDR significance, and split-half reliability is 0.935. The Network Comparison Test detects few edge differences between modalities (3-5\% of edges), while bridge symptoms diverge. For eating disorders, the item-level results are exploratory. Identified linguistic features replicate the same direction across all 3 corpora, and they cluster in three construct families.
\end{abstract}

\section{Introduction}

Network psychometrics treat mental disorders as patterns of interacting symptoms \citep{Borsboom2017, Robinaugh2020}, with bridge symptoms as candidate targets for transdiagnostic intervention \citep{Jones2021}. The framework has been productive but relies almost entirely on self-report questionnaires. \citet{Fried2017} documents 52 depression symptoms across seven scales with a mean overlap of 0.36; which nodes appear, and which emerge as bridges, depends on the symptoms the instrument uses.

To our knowledge, this is the first symptom-resolved psychometric network estimated from naturalistic text and the first comparison of clinical and linguistic symptom networks within the same instrument frame. A two-stage bi-encoder cascade scores each user against each questionnaire item, and the resulting per-user, per-symptom matrix feeds Gaussian Graphical Model estimation. We apply the framework across two eRisk corpora (depression as the main finding, eating disorders as exploratory), with additional analyses on a third (anorexia). The methodological contribution lies not in the retrieval architecture, adapted from existing eRisk systems, but in estimating item-resolved symptom networks from that output and comparing them formally to questionnaire-based networks via the Network Comparison Test. Symptoms surface in text either explicitly ("I haven't enjoyed anything in weeks") or through emotional tone and word choice; we use retrieval for the first and LIWC-22 \citep{Boyd2022} with pretrained emotion classifiers for the second.

We address two questions. \textbf{RQ1}: Do networks estimated from per-user, per-symptom scores produced by a retrieval pipeline for social media posts resemble questionnaire-based networks? \textbf{RQ2}: Are there linguistic features that separate symptomatic from non-symptomatic users consistently across disorders and recruitment frames?

Our contributions are: (1) a comparison of clinical and linguistic symptom networks using the Network Comparison Test, (2) a cross-disorder taxonomy of several replicating linguistic features, and (3) a reproduced state-of-the-art framework for symptom retrieval with an added first-person filter.

\section{Related Work}

In psychopathology network models, nodes are symptoms and edges are partial correlations after conditioning on all other symptoms. The standard estimator is the Gaussian Graphical Model (GGM) with EBIC-tuned graphical LASSO \citep{EpskampFried2018}, yielding edge weights, centrality (Expected Influence), community structure, and bridge symptoms \citep{Jones2021}; stability is assessed via case-dropping bootstraps and the CS-coefficient \citep{Epskamp2018bootnet}.

Bi-encoder retrieval \citep{karpukhin-etal-2020-dense} is a standard Information Retrieval (IR) architecture, applied to mental health symptom retrieval such as the eRisk shared task \citep{Parapar2024}. Elevated first-person pronoun use reliably correlates with depression severity \citep{Rude2004, Tackman2019}; lower analytic thinking with psychological distress \citep{CORBIN2023100645}; and eating-disorder communities show identifiable lexical profiles \citep{Chancellor2016}. Pretrained emotion classifiers \citep{demszky-etal-2020-goemotions, hartmann2022emotionenglish} produce post-level emotion distributions complementary to fixed lexicons; whether they add signal beyond LIWC in symptom-network analyses has not been tested until now.

\citet{perez-etal-2023-semantic} score symptoms via semantic similarity to BDI-II phrasings on the same eRisk corpora, focusing on questionnaire-total prediction and sentence labeling, but not per-symptom network estimation and cross-disorder replication.

\section{Data and Methodology}

Our framework has four components: We first extract linguistic and emotion features from each user’s posts (1), score each user on each questionnaire item by retrieving the posts closest in meaning to the item wording, which yields a continuous symptom expression score (2). We validate these scores against users’ questionnaire responses and test whether the linguistic patterns hold across disorders and datasets. We then use the scores in two analyses. The bridge analysis (3) correlates each symptom score with each linguistic feature. The network analysis (4) builds one symptom network from questionnaire responses and one from symptom expression scores, and compares their structure, including their communities.

\subsection{Datasets}

We use three eRisk corpora. In the \textbf{Depression} corpus (eRisk 2021, $N=169$), users completed the Beck Depression Inventory-II (BDI-II; \citealt{beck1996beck}), whose 21 items each cover one depressive symptom, scored 0-3. In the \textbf{Eating disorders} corpus (eRisk 2022-2025, $N=92$), users completed the Eating Disorder Examination Questionnaire (EDE-Q; \citealt{FairburnBeglin1994}); we use its 22 items scored 0-6, which form four subscales (Restraint, Eating Concern, Shape Concern, Weight Concern). The \textbf{Anorexia} corpus (eRisk 2024, $N=784$) has only binary labels based on self-stated diagnosis. In the two questionnaire corpora, per-item responses are the ground truth for validating symptom expression scores and for estimating clinical networks. All items and the short labels used in the paper are listed in Appendix~\ref{sec:appendix} (Tables~\ref{tab:bdi_mapping} and~\ref{tab:edeq_mapping}). Item meanings are fixed by the instruments: e.g., Crying is a depressive symptom, and its queries target that sense rather than a positive one.

The eRisk 2025 Task~1 release \citep{Parapar2025}, 28,025 annotated sentences pooled from earlier editions, is used separately to evaluate retrieval; given the similar data and task, we reproduce its top-performing systems. We do not compare against the official 2025 runs: our system was not in the judgment pool, so relevant sentences it retrieved but that were never judged would count as non-relevant, biasing the comparison against it.

\subsection{Symptom Expression Validation}

\subsubsection{Symptom Retrieval Pipeline} \label{sec:symptom_retrieval_pipeline}

We score each user against each questionnaire item using their own posts. Posts and items are embedded in a shared semantic space, and the cosine similarities are aggregated into an $N_\text{users} \times N_\text{items}$ matrix of \emph{symptom expression scores}, which mirrors a completed questionnaire. The scores occupy a narrow range (EDE-Q subscale medians 0.41-0.46), as is typical of dense retrieval over a domain-homogeneous corpus; since all downstream analyses use Spearman correlations, only the ordering of users matters.

Posts are first filtered to those with LIWC first-person singular density $\texttt{i} \geq 1.0$ \citep{PennebakerMehl2003, Tausczik2010}, so that scores reflect the user's own experience; an ablation explores this filter.

Retrieval follows the two-stage bi-encoder cascade of the top-performing eRisk 2025 systems. Stage~1 ranks filtered posts with \texttt{all-MiniLM-L6-v2}\footnote{\url{https://huggingface.co/sentence-transformers/all-MiniLM-L6-v2}} and keeps the top 1,000 per symptom; Stage~2 re-encodes them with \texttt{bge-large}\footnote{\url{https://huggingface.co/BAAI/bge-large-en-v1.5}}, and the user score is their mean Stage-2 similarity. Since no user has more than 1,000 filtered posts (maximum 878), Stage~1 keeps all of them, and the score averages over a user's full self-referential history; Stage~1 selection matters for sentence retrieval on eRisk. BDI-II queries average the embeddings of severity phrasings 1-3 (severity 0 excluded) and are L2-renormalized; EDE-Q items have a single phrasing each.

Sentence-level relevance data exist only for depression, so we select the configuration there and apply it to EDE-Q unchanged; tuning it on ED would mean optimizing against the same responses used for validation. As the encoders are general-purpose and only the queries change, we treat the ED analysis as a test of transfer across instruments. Transfer is partial: depression subsampled to $N{=}92$ (2,000 draws) keeps a median composite $\rho$ of 0.43 and 16/21 FDR-significant items and never falls to the ED level, so sample size alone might not explain the weaker ED results, which we report as exploratory.

As classical baselines, we score users with BM25 and TF-IDF (corpus-level term weights), keeping the post pool, filter, and aggregation fixed so that only the scoring method differs. To test sensitivity, we recompute the depression scores under alternative query recipes, aggregations (median, maximum, top-$k$ mean), candidate pools ($k{=}10$-$250$), and encoders (MiniLM only), and re-estimate the network for each.

\subsubsection{Linguistic Feature Extraction}

Linguistic features capture how distress manifests in word choice, emotional tone, and topic distribution without the user naming a symptom directly. We use three feature families that have established psychometric and clinical histories: \textit{lexicon-based features} using LIWC-22 \citep{Boyd2022} for functional, affective, cognitive, social, and content-domain dimensions; \textit{corpus-based emotion features} using the Hartmann 7-emotion DistilRoBERTa classifier \citep{hartmann2022emotionenglish} and GoEmotions 28-emotion classifier \citep{demszky-etal-2020-goemotions} for emotion-probability distributions; and \textit{within-user variability indices} (standard deviations of emotion probabilities across the user's posts). Post-level features are averaged to user level. The full feature vector contains 191 dimensions. 

\subsubsection{Dimensional Validation}

The validation question is whether NLP-derived symptom expression scores actually correspond to clinical responses on the same items. In this sense, for each user we compute a per-symptom symptom expression score and a composite symptom expression score as the mean across symptoms. The validation targets are the user's actual questionnaire responses on the corresponding items, and the questionnaire total for the composite. Validation correlations use Spearman $\rho$ and per-symptom correlations are FDR-corrected via Benjamini-Hochberg \citep{Benjamini1995} within the item family for each instrument. Bootstrap 95\% confidence intervals on each $\rho$ use 1000 resamples. We interpret $|\rho|<0.20$ as negligible, $0.20$-$0.30$ as weak, $0.30$-$0.50$ as adequate, and $\geq 0.50$ as strong, following \citet{GignacSzodorai2016}; Cohen's benchmarks \citep{Cohen1988} are not used here because they were not derived from an empirical distribution of psychological correlations.

At $N{=}92$ with 22 items, per-item ED analyses have limited power, and the power-matched comparison (Section \ref{sec:symptom_retrieval_pipeline}) shows that the weaker ED results are not explained by sample size alone. ED per-item results are therefore reported as exploratory.

\subsubsection{Split-half Reliability}

To assess the internal consistency of scores independent of their convergence with self-report, each user's filtered posts are randomly split into two halves of equal size (within-user). Per-symptom symptom expression scores are recomputed on each half using the same retrieval pipeline. Between-half correlations across users (Spearman $\rho$) are then Spearman-Brown corrected to estimate full-length reliability, with literature supporting thresholds of above 0.7 as acceptable for exploratory research, above 0.8 as good, and above 0.9 as most preferable \citep{Nunnally1994}.

\subsubsection{Self-attribution Filter Ablation}

The LIWC \texttt{i}-filter is a lexical measure for self-attribution. We test whether it carries real measurement weight by comparing with applying no filter (all posts retained).

\subsubsection{eRisk 2025 Retrieval Evaluation}

We also evaluate against the eRisk 2025 Task~1 benchmark \citep{Parapar2025}, which uses pooled human relevance judgments at the sentence level. The release provides a test corpus of ${\approx}17.5$M sentences from 9,000 users with majority-vote (2/3 assessors) relevance judgments, comprising ${\approx}$21K judged sentence-symptom pairs across the 21 BDI-II items; we also tested a cross-encoder with fine-tuning using the corresponding training qrels.

The primary metric we opt for is bpref (binary preference) \citep{Buckley2004}, an evaluation metric for the relative ranks of relevant and irrelevant documents which is robust to incomplete relevance judgments arising from shallow pooling. Auxiliary metrics (MAP, nDCG@10/100/1000, P@10, MRR) are reported but biased downward under the same condition. Multiple configurations are compared: BGE bi-encoder only; BGE + off-the-shelf cross-encoder; two-stage cascade (primary); MiniLM bi-only baseline; BGE + fine-tuned cross-encoder.

\subsection{Cross-disorder and Cross-dataset Analyses}

The cross-disorder analysis asks whether the same linguistic features separate symptomatic from control users across the three corpora, which differ in disorder and recruitment. For each corpus, we compare positive and control users on 134 linguistic features with Mann-Whitney $U$ tests, reporting rank-biserial $r$ with BH-FDR correction across features. Each feature is categorized by its pattern of significance and direction: \textit{transdiagnostic} (significant in all three corpora, same direction), \textit{shared} (significant in exactly two, same direction), \textit{disorder-specific} (significant in one, reported per disorder), \textit{mixed-direction} (significant in two or more, with opposite signs), and \textit{non-discriminating} (significant in none). To test whether the transdiagnostic features form construct families, we represent each by its three effect sizes, standardized per corpus, and group them with Ward hierarchical clustering, choosing the number of families by silhouette score over $k\in\{3,4,5\}$.

The cross-dataset analysis applies the EDE-Q pipeline unchanged to the anorexia corpus and compares symptom expression scores between anorexia-positive and control users, and between anorexia-positive and ED-positive users. As the anorexia corpus has no EDE-Q responses, this tests whether the scores generalize to an independently recruited sample, not their dimensional validity.

\subsection{Network Analysis} 

The retrieval pipeline produces a per-user, per-symptom score matrix that mirrors a completed questionnaire's structure. From that matrix we estimate a symptom network and compare it to the network estimated from the corresponding clinical questionnaire responses. The networks are compared at four structural levels: edge weights, centrality, community membership, and bridge symptoms.

\subsubsection{Network estimation}

Networks are estimated as Gaussian Graphical Models via EBICglasso \citep{Friedman2008glasso} with $\gamma$ (which controls model complexity and sparsity) $= 0.5$ \citep{Foygel2010}. All networks use Spearman correlations as input: this accommodates the ordinal nature of questionnaire responses, is robust to outliers in NLP cosine scores, and gives a uniform estimation procedure across clinical and linguistic networks, which is a prerequisite for the Network Comparison Test \citep{VanBorkulo2022}.

\subsubsection{Network metrics}

 We use Expected Influence \citep{Robinaugh2016} as it handles negative partial correlations correctly. For bridge centrality we employed one-step Bridge Expected Influence \citep{Jones2021}, computed with each network's own Walktrap partition. Bridge centrality depends on community membership; we quantify cross-modality bridge agreement by Spearman rank correlation of the Bridge EI vectors between the clinical and NLP symptom expression networks, a standard centrality-stability measure \citep{Epskamp2018bootnet}.

\subsubsection{Community Detection}
\label{community_meth}

Walktrap algorithm \citep{Pons2005} at $t=4$ is applied directly to the EBICglasso adjacency matrix. Partition agreement between clinical and NLP communities is measured by Adjusted Rand Index \citep{Hubert1985}, which looks at the similarity between two data clusterings while correcting for random chance. While classic cluster recovery simulation benchmarks strictly demand an ARI > 0.65 \citep{Steinley2004}, such thresholds assume identical underlying data structures. Following the chance-corrected classification framework of \citet{Landis1977}, we defined an ARI > 0.25 as fair cross-modality structural alignment and an ARI > 0.50 as moderate-to-strong alignment.

\subsubsection{Network Comparison Test and Stability}

The NCT \citep{VanBorkulo2022} provides permutation-based hypothesis tests for differences between two estimated networks. With 1000 permutations we compute the M-test (maximum edge difference), global strength test, per-edge difference tests with BH-FDR correction across edges, and per-node centrality difference tests with BH-FDR correction across nodes. 

CS-coefficients are estimated via case-dropping bootstrap. The CS-coefficient is the maximum case-drop proportion that preserves correlation >= 0.7 with full-sample centrality in 95\% of resamples. netSimulator \citep{EpskampFried2018} provides power analysis at observed N by simulating data under a known network and estimating recovered sensitivity, specificity, and edge-weight correlation.

\subsubsection{Symptom-feature bridge matrices}

For each symptom and linguistic feature, we compute Spearman $\rho$ between user-level symptom expression scores and user-level feature averages, with BH-FDR correction over all symptom-feature pairs in a matrix. We compute three matrices: BDI items $\times$ features ($N{=}169$; $21\times191{=}4011$ tests), EDE-Q items $\times$ features in the ED corpus ($N{=}92$; 4202 tests), and the same EDE-Q items $\times$ features in the anorexia corpus ($N{=}782$; 4180 tests).

Since the ED matrix is underpowered, we identify associations that hold across both eating-disorder corpora: pairs that are FDR-significant in the anorexia corpus and reach $|\rho|\geq0.20$ in the same direction in the ED corpus. Replication in two independently recruited samples makes these pairs less likely to reflect corpus-specific artefacts. The 0.20 floor was pre-specified as the smallest effect size of interest \citep{Lakens2018}; at $N{=}92$ it gives about 48\% power, above $r{=}0.16$, the effect detectable with 33\% power at this sample size \citep{Simonsohn2015}. Results at floors of 0.15 and 0.25 are reported in Appendix~\ref{sec:appendix}, Table~\ref{tab:sensitivity}.

LIWC \texttt{i} serves both as the post-level filter and as a feature in the bridge matrices. The two are related but distinct: the filter is a binary gate on individual posts, while the feature is a continuous average over all of a user's posts.

\section{Results}

\subsection{Symptom Expression Retrieval}

\subsubsection{eRisk 2025 Task 1 Annotated Data Performance}

Performance metrics computed on the eRisk 2025 Task 1 annotated data show that the cascade approach achieves the highest bpref (0.6746). Fine-tuning the cross-encoder on eRisk training qrels does not improve over off-the-shelf weights ($\Delta\text{bpref}$=0.0017). The cascade no-filter approach matches the primary on every metric, confirming that the \texttt{i}-filter affects only user-level aggregation, not sentence ranking. Full results are visible in Appendix \ref{sec:appendix}, Table \ref{tab:erisk}.

\subsubsection{Dimensional Validation}

The primary configuration outperforms the alternative configurations and the sparse baselines (Appendix~\ref{sec:appendix}, Table~\ref{tab:retrieval}): 20 of 21 BDI items reach FDR significance, and the composite correlation is $\rho{=}0.434$ ($p{<}10^{-8}$). Q17 (Irritability) does not survive FDR correction, and Q3 (Past Failure) and Q6 (Punishment) are the weakest significant items.

ED results are weaker. The composite correlation is $\rho{=}0.200$ ($p{=}0.057$), and no item reaches FDR significance. Since depression subsampled to $N{=}92$ never falls to this level (Section~\ref{sec:symptom_retrieval_pipeline}), the gap is not only a matter of sample size; it may reflect weaker transfer of the configuration, less explicit textual expression of EDE-Q symptoms, or differences between the instruments. Subscale correlations range from 0.15 to 0.20, with Shape Concern ($\rho{=}0.202$, $p{=}0.055$) and Eating Concern ($\rho{=}0.200$, $p{=}0.057$) closest to significance.

\subsubsection{Split-half Reliability}

Within-user split-half reliability is measured through composite Spearman-Brown corrected, resulting in a value of 0.935. All 21 BDI symptoms exceed 0.90 (range 0.901-0.944), indicating that symptom expression scores are highly stable across random post halves. The moderate composite correlation with self-report ($\rho$=0.434) reflects more between-measure differences rather than unreliability in the NLP measurement.

\subsubsection{Sensitivity to Retrieval Choices}

Validity is stable across query recipes (severity 1, 3, or averaged: $\rho{=}0.43$-$0.45$, 19-20 items FDR-significant), between mean and median aggregation ($\rho{=}0.43$ vs.\ $0.41$), and largely with a weaker encoder (MiniLM only: $\rho{=}0.36$, 16 items). It depends on averaging over the full post history, with scoring only the top 10-50 posts per symptom reducing $\rho$ to 0.19-0.22, and maximum aggregation to 0.16. The gain of the primary configuration over the BGE-only and cross-encoder configurations, which score users by their top 20 posts, therefore comes from this aggregation. Across variants, the re-estimated NLP networks correlate 0.68-0.97 with the primary network.

\subsubsection{Self-attribution Filter Ablation}

The LIWC filter and no-filter conditions are empirically equivalent ($\Delta \rho$ = 0.009). The filter is therefore not needed for validity; we retain it for theoretical grounding, since the score should reflect the user's own state rather than their commentary on others.

\subsection{Cross-disorder and Cross-dataset Analyses}

Applied to the anorexia corpus, the EDE-Q pipeline yields $782 \times 22$ symptom scores (2 of 784 users lack sufficient posts after filtering). All 22 items separate anorexia-positive from control users (rank-biserial $|r|{=}0.91$-$0.94$, all FDR-significant), consistent with the pipeline retrieving topically appropriate content. ED users score higher than anorexia users on all four subscales (e.g., Restraint medians 0.441 vs.\ 0.414; $r{\approx}0.44$-$0.47$, all $q{<}0.001$), consistent with questionnaire-screened ED users writing more EDE-Q-relevant content than anorexia users, whose histories cover a wider range of topics.

\begin{figure}[ht]
\centering
\includegraphics[width=0.99\columnwidth]{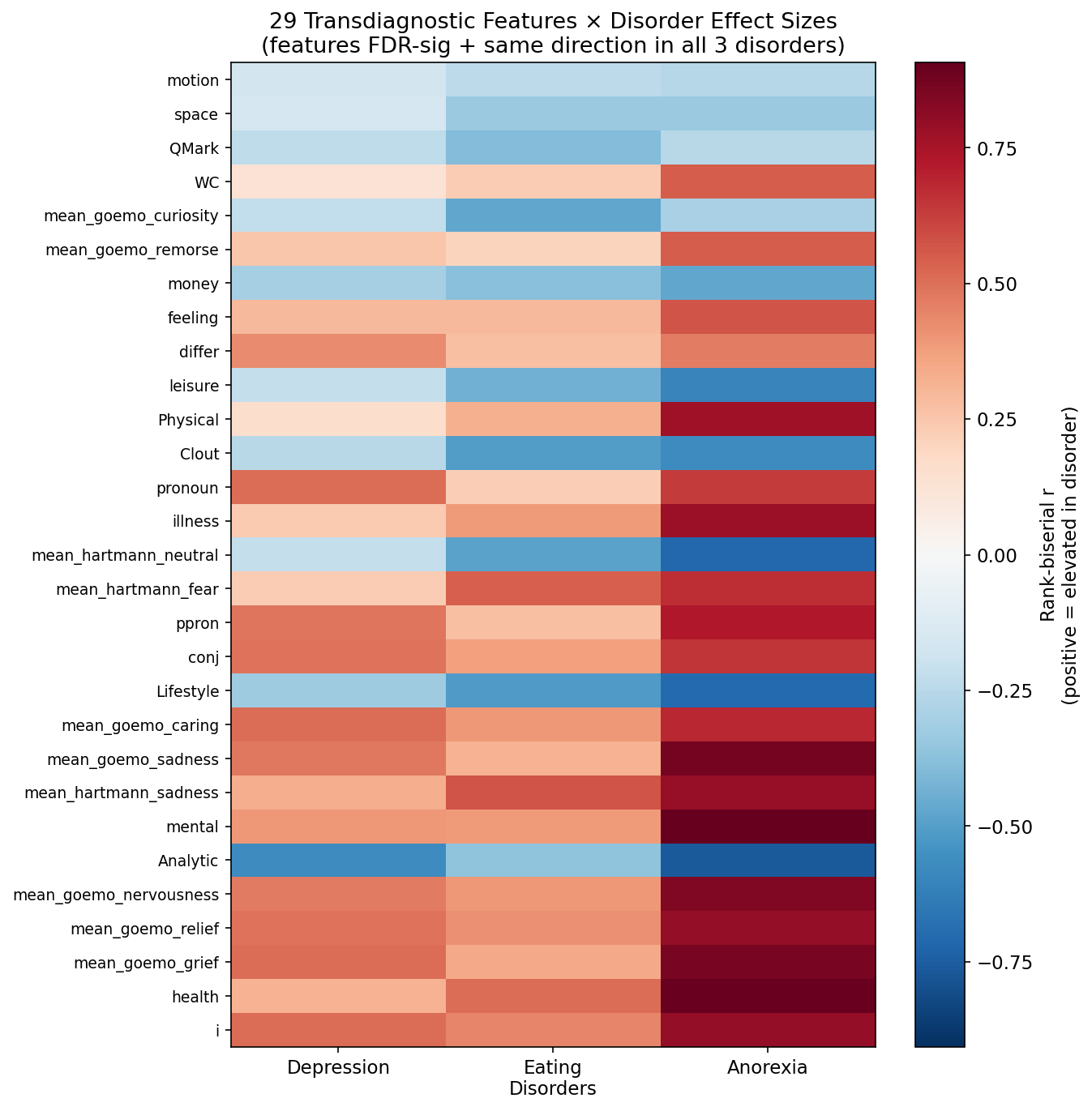}
\caption{Signed rank-biserial effects of the 29 transdiagnostic features in each disorder. Positive values indicate features elevated in symptomatic users.}
\label{fig:transdiag}
\end{figure}

\begin{figure}[ht] 
\centering
\includegraphics[width=\columnwidth]{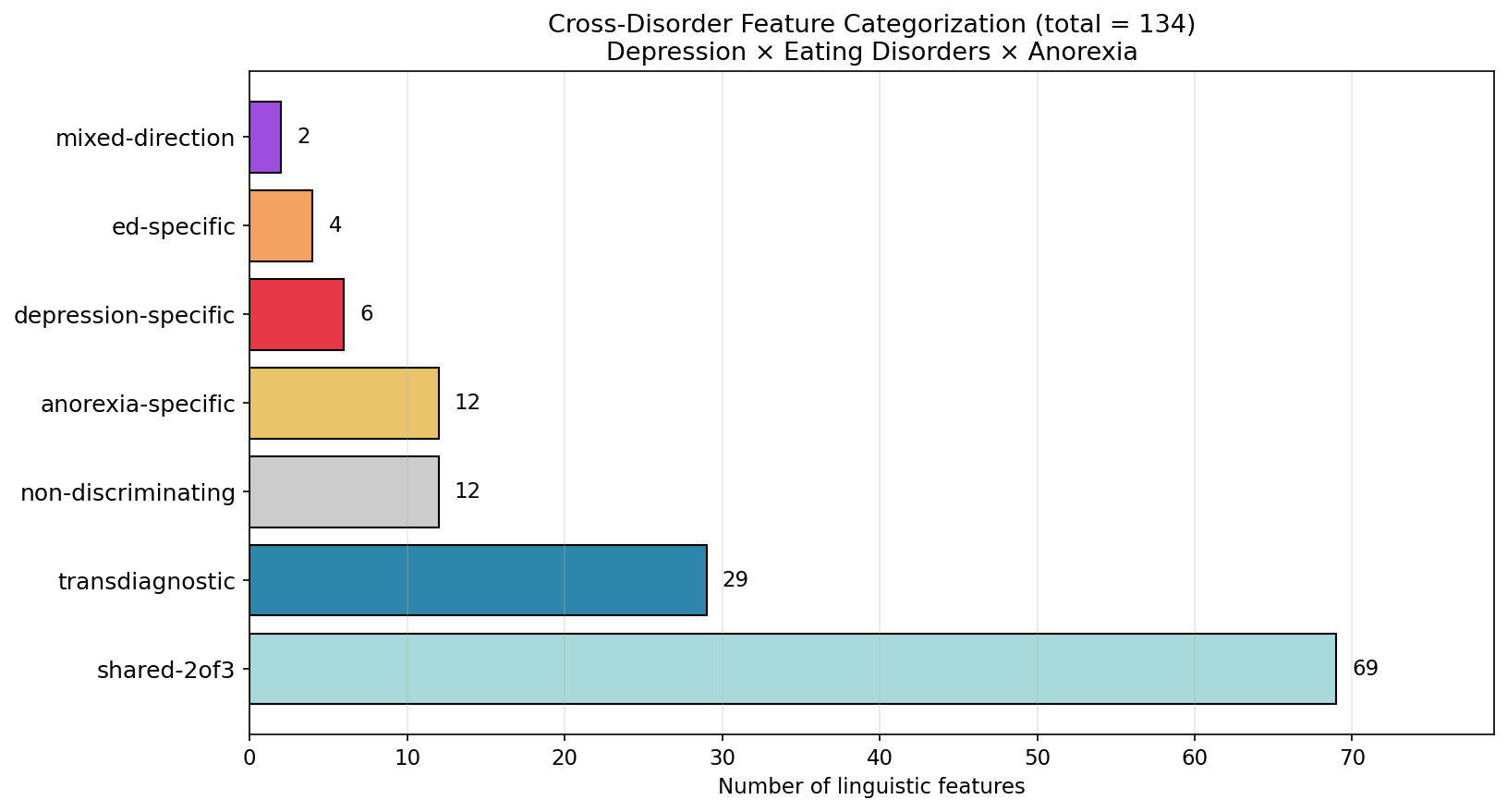} 
\caption{Distributions of features across the established categories in the cross-disorder comparison.}
\label{tab:cross_dis_categ}
\end{figure}

Of the 134 features, 29 are transdiagnostic (Figure~\ref{fig:transdiag}); the rest are shared by two disorders (69), disorder-specific (22: anorexia 12, depression 6, ED 4), non-discriminating (12), or mixed-direction (2) (Appendix~\ref{sec:appendix}, Figure~\ref{tab:cross_dis_categ}). The strongest is the first-person singular pronoun (LIWC \texttt{i}; mean $r{=}+0.58$; anorexia $+0.79$, depression $+0.50$, ED $+0.45$), which extends the depression finding of \citet{Rude2004} to two further disorders and supports reading it as a marker of self-focused distress rather than of depression specifically. Since all 29 features keep their direction across three independently recruited samples, and the two emotion classifiers, trained on different data and label sets, agree in direction, the pattern is unlikely to be an artefact of a single study design, recruitment frame, or instrument.

Ward clustering groups the transdiagnostic features into three families (silhouette $=0.55$; Appendix~\ref{sec:appendix}, Table~\ref{tab:families}): (1) \emph{reduced externally oriented language} (analytic style, clout, neutral tone, curiosity, and references to money, space, motion, and leisure; mean $r{=}-0.39$); (2) \emph{elevated self-focused affect} (first-person pronouns with GoEmotions grief, sadness, nervousness, caring, and relief; mean $r{=}+0.53$); and (3) \emph{elevated somatic content} (LIWC health, illness, and physical terms with Hartmann fear and sadness; mean $r{=}+0.44$). The set is therefore not reducible to a single negative-affect dimension, and it includes positive-valence emotions (relief, caring) that are elevated in symptomatic users.

\subsection{Network structure: Clinical versus Linguistic} \label{sec:net_struc}
\begin{figure*}[t]
  \centering
  \begin{subfigure}[b]{0.49\textwidth}
    \centering
    \includegraphics[width=\textwidth, pagebox=cropbox]{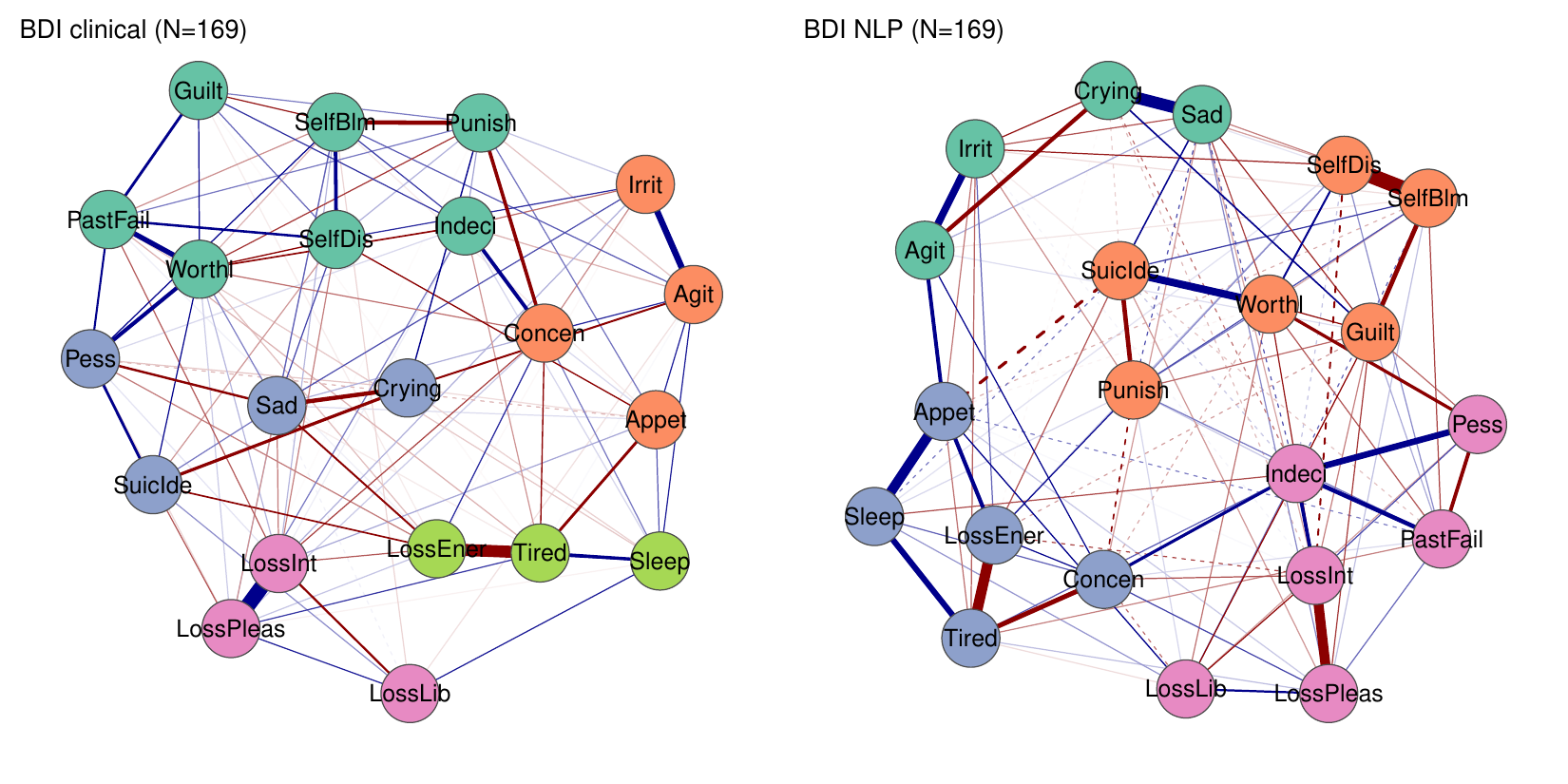}
    \caption{BDI-II symptom networks ($N=169$).}
    \label{fig:bdi_networks_sub}
  \end{subfigure}
  \hfill
  \begin{subfigure}[b]{0.49\textwidth}
    \centering
    \includegraphics[width=\textwidth, pagebox=cropbox]{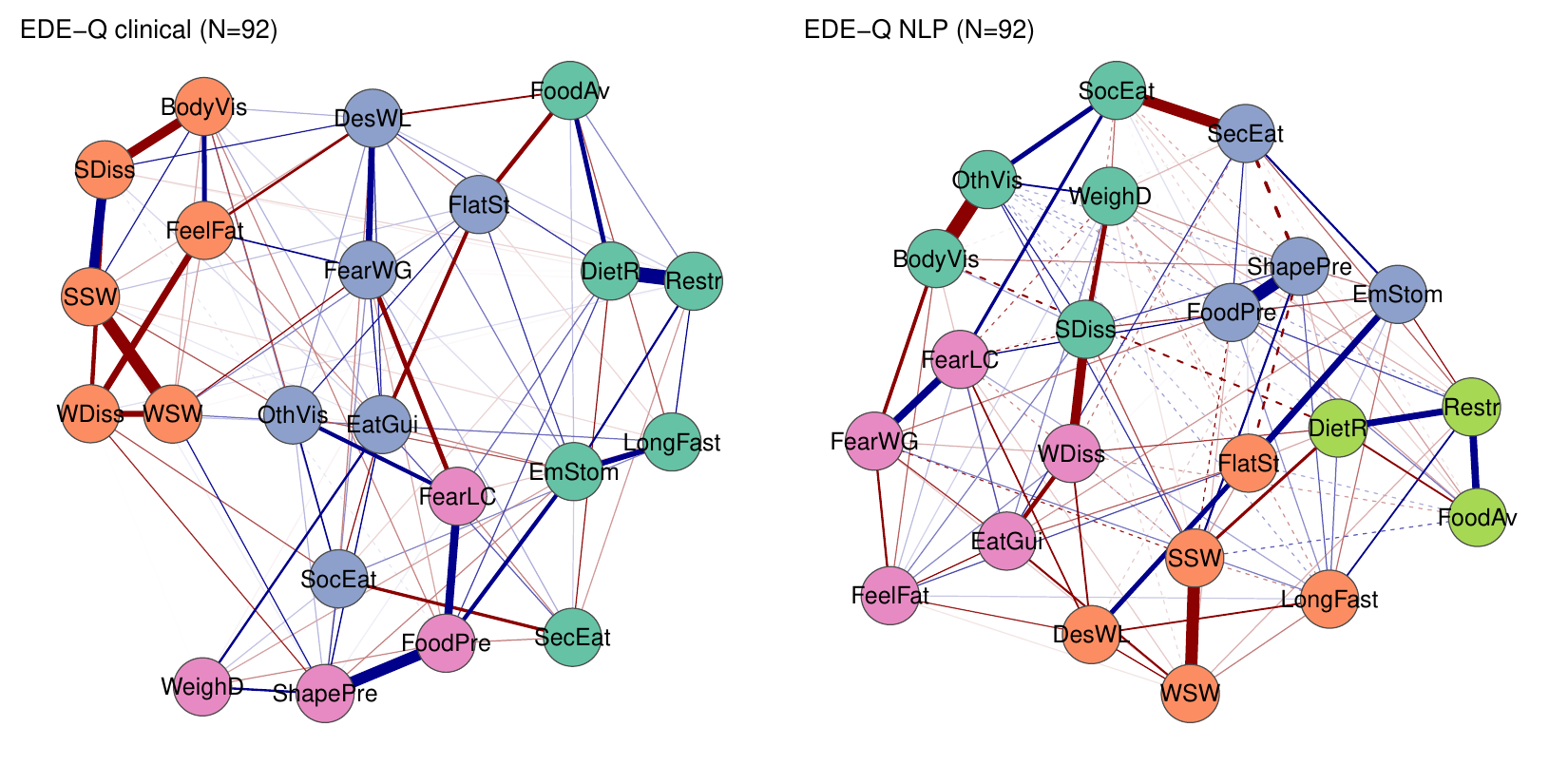}
    \caption{EDE-Q item-level symptom networks ($N=92$).}
    \label{fig:edeq_item_networks_sub}
  \end{subfigure}
  \caption{Symptom networks estimated with EBICglasso ($\gamma=0.5$, Spearman correlations). Nodes are colored by each network's Walktrap partition; blue edges are positive partial correlations, red dashed edges negative ones. Node abbreviations are listed in Appendix~\ref{sec:appendix}, Subsection~\ref{abrev_labels}.}
  \label{fig:combined_networks}
\end{figure*}

Both language-derived networks are denser and have higher global strength than the clinical ones (Appendix~\ref{sec:appendix}, Table~\ref{tab:networks_global}), since symptom expression scores share more variance across items than questionnaire responses; cross-modality comparisons therefore rely on per-edge tests rather than global strength.

\begin{table}[ht]
\centering
\caption{Top-5 Expected Influence (EI) nodes.}
\label{tab:ei}
\resizebox{0.75\columnwidth}{!}{%
\begin{tabular}{clc lc}
\toprule
 & \multicolumn{2}{c}{\textbf{Clinical Network}} & \multicolumn{2}{c}{\textbf{NLP Network}} \\
\cmidrule(r){2-3} \cmidrule(l){4-5}
\textbf{Rk} & \textbf{Node} & \textbf{EI} & \textbf{Node} & \textbf{EI} \\
\midrule
\multicolumn{5}{l}{\textit{BDI (Overlap: 2/5)}} \\
1 & Worthlessness & 1.23 & Tiredness & 1.34 \\
2 & Loss Interest & 1.16 & Worthlessness & 1.31 \\
3 & Loss Energy & 1.13 & Guilt & 1.15 \\
4 & Tiredness & 1.06 & Self Blame & 1.13 \\
5 & Self Blame & 1.03 & Loss Pleasure & 1.10 \\
\hline
\multicolumn{5}{l}{\textit{EDE-Q (Overlap: 3/5)}} \\
1 & Desire Weight Loss & 1.22 & Restraint & 1.35 \\
2 & Food Preoccupation & 1.16 & Eating Guilt & 1.31 \\
3 & Eating Guilt & 1.13 & Weight Self Worth & 1.15 \\
4 & Weight Self Worth & 1.11 & Desire Weight Loss & 1.13 \\
5 & Shape Preoccupation & 1.10 & Empty Stomach & 1.11 \\
\bottomrule
\end{tabular}%
}
\end{table}

The most influential nodes are largely shared across modalities (Table~\ref{tab:ei}): Worthlessness and Self Blame for BDI, and Eating Guilt, Weight Self Worth, and Desire Weight Loss for EDE-Q. These are self-directed cognitions that are often stated explicitly. Restraint ranks first in the ED language network, consistent with how often restriction and fasting are described in such posts.

\begin{table}[ht]
\centering
\caption{Community detection (Walktrap, $t=4$) and ARI between clinical and NLP partitions.}
\label{tab:community}
\resizebox{0.85\columnwidth}{!}{%
\begin{tabular}{lcccc}
\toprule
Network & Communities (clinical) & Communities (NLP) & ARI \\
\midrule
BDI 21-item      & 5 & 4 & 0.337 \\
EDE-Q 22-item    & 4 & 5 & 0.066 \\
EDE-Q 4-subscale & 2 & 1 & 0.000 \\
\bottomrule
\end{tabular}%
}
\end{table}

Community structure is only partly shared (Table~\ref{tab:community}). The NLP subscale network places all four EDE-Q subscales in a single community, suggesting that ED-related language does not separate Restraint, Eating, Shape, and Weight Concern as the instrument's subscales propose. Against the two-factor BDI-II solution of \citet{beck1996beck} (Somatic-Affective, Cognitive), both networks recover the structure above a permutation null (ARI $=0.21$ clinical, $0.25$ linguistic; $p=0.005$ and $0.003$). The linguistic network is about as close to the factor solution as to the clinical network (ARI $=0.34$), so the two are related but distinct.

Bridge centrality diverges more than Expected Influence (Appendix~\ref{sec:appendix}, Table~\ref{tab:bridges}). The BDI clinical and NLP top-5 bridges do not overlap, and for EDE-Q only Empty Stomach is shared. Across the full rankings, Bridge EI vectors correlate at $\rho{=}-0.15$ for BDI and $\rho{=}+0.10$ for EDE-Q. Since bridge scores depend on community membership, and community detection is sensitive to small changes near community boundaries \citep{Fortunato2010, Good2010}, the partial community agreement (ARI $=0.337$ and $0.066$) carries into these rankings.

The BDI NLP bridges are mostly self-directed cognitions (Guilt, Worthlessness) and high-arousal states (Agitation, Irritability), whereas the clinical bridges include cognitive-motivational (Concentration, Loss of Interest) and somatic (Loss of Energy) symptoms, which may be endorsed on questionnaires more than they are written about. Both modalities identify an anhedonia bridge but differ in its form: Loss of Interest clinically, Loss of Pleasure linguistically. The self-referential emphasis of the NLP bridges is consistent with \citet{Rude2004} and \citet{Tackman2019}.

\subsection{Network Comparison Test}

For BDI, 6 of 210 edges (2.9\%) differ FDR-significantly between modalities; for EDE-Q items, 12 of 231 (5.2\%); for EDE-Q subscales, none. This agreement is expected, since both networks are estimated over the same items and inherit their shared content, and it indicates that the text-derived network preserves the instrument's structure, consistent with both networks recovering the BDI-II factor solution (Section~\ref{sec:net_struc}). Edge agreement is therefore necessary but not sufficient, and it holds across all retrieval variants in the sensitivity analysis (Appendix~\ref{sec:appendix}, Table~\ref{tab:sensitivity_retrieval}). The more informative comparisons are at the level of bridges, where the modalities diverge, and of individual nodes.

Two nodes differ in centrality: Guilt is more central clinically ($\Delta\text{EI}{=}-0.544$, $q{=}0.021$), and Shape Self Worth more central linguistically ($\Delta\text{EI}{=}+0.395$, $q{=}0.044$). This suggests that guilt is endorsed on the questionnaire more than it is written about, while shape-based self-evaluation is written about beyond its questionnaire weight, a difference between what people put into words and what they report on a scale.

To test whether the scores carry symptom-specific information beyond overall severity, we ran a multitrait-multimethod analysis after removing each user's mean score from both the text and questionnaire profiles (ipsatizing), so that only relative symptom elevation remains. Same-symptom correlations between text and questionnaire stay positive (mean $r{=}0.08$), while different-symptom correlations fall to zero (mean $r{=}0.00$). A supervised baseline we also tested shows no such separation. The effect, while modest, indicates that the retrieval scores capture which symptoms are elevated within a user, not only how severe they are.

\subsection{Network Stability}

BDI clinical CS scores 0.515, supporting the centrality interpretation, while EDE-Q clinical CS being at 0.359 is in the sufficient range, warranting caution for interpretation. Bootstrap edge CIs (1000 resamples) show the subset of strongest edges with intervals excluding zero, with BDI showing tighter intervals than EDE-Q. netSimulator for the EDE-Q clinical network at $N=92$ yields edge sensitivity of 0.65, specificity of 0.67, and edge-weight correlation of 0.71. The EDE-Q clinical network is recoverable in coarse structure but not at the level of individual edges.

\subsection{Symptom-feature bridge matrices}

\begin{table}[ht]
\centering
\caption{Symptom-feature bridge matrices. Strong: $q < 0.05$ and $|\rho| \geq 0.20$. $\dagger$Descriptive only at $N=92$. $\ddagger$242 pairs also reach $|\rho_\text{ED}| \geq 0.20$ in the same direction (cross-population intersection).}
\label{tab:bridge_matrix}
\resizebox{0.99\columnwidth}{!}{%
\begin{tabular}{llccccc}
\toprule
Network & $N$ & Tests & FDR-sig & \% & Strong & \% \\
\midrule
BDI $\times$ features                  & 169 & 4011 &  782 & 19.5 & -   & -   \\
EDE-Q $\times$ features (ED)$\dagger$  &  92 & 4202 &    0 &  0.0 & -   & -   \\
EDE-Q $\times$ features (anorexia)$\ddagger$ & 782 & 4180 & 3678 & 88.0 & 2602 & 62.2 \\
\midrule
\multicolumn{7}{l}{\small By feature category (anorexia matrix):} \\
\quad LIWC       & - & 3366 & 2965 & 88.1 & 2041 & 60.6 \\
\quad GoEmotions & - &  616 &  545 & 88.5 &  451 & 73.2 \\
\quad Hartmann   & - &  154 &  131 & 85.1 &   90 & 58.4 \\
\quad Variability& - &   44 &   37 & 84.1 &   20 & 45.5 \\
\bottomrule
\end{tabular}%
}
\end{table}

The anorexia matrix is dense: 88.0\% of pairs are FDR-significant and 62.2\% strong ($|\rho|\geq0.20$; Table~\ref{tab:bridge_matrix}). For ED, no pair survives FDR correction over 4202 tests at $N{=}92$, so ED inference rests on the cross-population intersection, the FDR-significant anorexia pairs that also reach $|\rho_\text{ED}|\geq0.20$ in the same direction in the ED corpus: 242 pairs (8.4\% of the 2885 anorexia FDR-significant pairs in the aligned $22\times153$ space).

For BDI ($N{=}169$), 21.4\% of emotion-feature pairs are FDR-significant, against 18.4\% of LIWC pairs. Grief is associated with Sadness and Tiredness ($\rho{\approx}+0.37$), and nervousness with Pessimism, Suicidal Ideation, Worthlessness, and Concentration ($\rho{\approx}+0.27$-$0.34$). This split is consistent with the distinction between melancholic and anxious depression \citep{Musil2018}, suggesting separable pathways rather than a single negative-affect dimension.

Empty Stomach recurs in three analyses, the EDE-Q clinical and NLP Bridge EI top-5 lists and the cross-population intersection, making it the strongest single bridge symptom the data support.

Within the anorexia corpus, emotion features are more often strongly associated with symptoms than LIWC features (strong pairs: GoEmotions 73.2\%, LIWC 60.6\%). In the cross-population set, each family's share roughly follows its number of features, so the data do not show one family generalizing better than the other.

\section{Conclusions and Future Work}

Symptom networks don't need to be built from questionnaires alone. To our knowledge, this is the first work to estimate symptom-resolved psychometric networks from social media text, compare them with questionnaire-based networks using the Network Comparison Test, and identify linguistic markers that replicate across three independently recruited corpora spanning two disorders. For depression, scores derived only from users' posts, using an adapted eRisk retrieval pipeline, correlate significantly with 20 of 21 BDI-II items (composite $\rho{=}0.434$), robustly across query recipes and encoders; for eating disorders, results are weaker and exploratory.

The comparison yields one central finding: text and questionnaires agree on which symptoms go together, but not on which ones connect the network. Most edges are shared, yet the bridge symptoms differ almost entirely, with language emphasizing self-directed cognitions such as guilt and worthlessness. Across disorders, the 29 replicating features can be clustered into three families: less externally oriented language, more self-focused affect, and more somatic content.

Future work will separate symptom endorsement from discussion through stance classification, model within-person change over time, and test transfer to other instruments and platforms.

\section{Limitations}

Each user receives one score against one questionnaire administration. Within-person dynamics are not captured; a user's symptom profile could shift substantially across their posting history, and the pipeline treats this variation as noise rather than signal.
All data come from Reddit. Generalization to other platforms or to clinical populations remains untested. This occurs because, to our knowledge, the eRisk collections are the only public corpora pairing item-level questionnaire responses with social media histories, which validating a text-derived network against a questionnaire-derived one requires.

We do not suggest that language-derived bridge nodes can directly guide clinical intervention without rigorous external validation. Bridge centrality here is a structural property of a text-derived network, not a demonstrated treatment target.

At $N{=}92$ with 22 items and four subscales, the ED corpus is underpowered for per-item and per-edge inference; this is quantified via netSimulator and is why ED per-item results are reported as exploratory. Beyond power, the EDE-Q presents a structural challenge that the BDI-II does not, its items span two distinct modalities: behavioral (fasting frequency, diet rules) and cognitive (shape preoccupation, fear of weight gain). Retrieval against a single item phrasing may conflate users who engage in the behavior with users who ruminate about it, conflating distinct symptom mechanisms within one score. A power-matched comparison further indicates that sample size alone does not account for the weaker ED results.

The anorexia corpus carries binary self-mention of diagnosis-derived labels, not dimensional questionnaire scores.

The pipeline operates on English text only; multilingual generalization is untested.

Self-attribution is operationalized via the LIWC \texttt{i} filter, which has not been validated against human annotations of post-level perspective.

Bi-encoder weights are used off-the-shelf. Cross-encoder fine-tuning was evaluated and did not improve performance; bi-encoder fine-tuning at scale was not evaluated and could in principle yield gains.

\section*{Ethical Considerations}

The eRisk corpora are publicly released through the shared task series under data-use agreements that anonymize users. We make no diagnostic or screening claims about specific individuals. Our analyses operate at the population level on aggregated linguistic signal.
Population-level patterns can be misappropriated for individual prediction; the dimensional $\rho$ values reported here are insufficient for individual diagnostic inference. Social media discourse is observational and confounded by platform-specific norms (subreddit topic, posting culture, anonymity); cross-platform generalization should not be assumed. Models that infer mental health states from text raise privacy concerns even when individuals are pseudonymous, and downstream deployment requires careful consideration of user consent and clinical oversight beyond the scope of methodological research.
The applications we consider beneficial are intervention research targeting bridge symptoms, transdiagnostic mechanism discovery, and methodological infrastructure for descriptive psychopathology research using large-scale text data.

\bibliography{custom}
\newpage

\appendix
\section{Appendix}
\label{sec:appendix}

\begin{table}[ht]
\centering
\caption{eRisk 2025 Task 1 results. \textit{bpref} \citep{Buckley2004} is the primary metric, robust to shallow pooling.}
\label{tab:erisk}
\resizebox{\columnwidth}{!}{
\begin{tabular}{llccccc}
\toprule
ID & Configuration & bpref & MAP & nDCG@10 & P@10 & MRR \\
\midrule
A  & BGE bi-encoder only              & 0.6515 & 0.5210 & 0.9884 & 0.9857 & 1.000 \\
B  & BGE + cross-encoder (off-shelf)  & 0.6506 & 0.5121 & 0.9461 & 0.9524 & 0.952 \\
\rowcolor{lgray}
C  & Two-stage cascade (primary)      & \textbf{0.6746} & 0.5333 & \textbf{0.9884} & \textbf{0.9857} & \textbf{1.000} \\
D  & MiniLM bi-only (baseline)        & 0.6713 & \textbf{0.5812} & 0.9699 & 0.9714 & 1.000 \\
E  & BGE + cross-encoder (fine-tuned) & 0.6523 & 0.5213 & 0.9571 & 0.9476 & 1.000 \\
F1 & Two-stage, no \texttt{i}-filter  & 0.6746 & 0.5333 & 0.9884 & 0.9857 & 1.000 \\
\bottomrule
\end{tabular}%
}
\end{table}

\begin{table}[ht]
\centering
\caption{Complete list of the three transdiagnostic construct families, with signed rank-biserial effect sizes per disorder (depression, eating disorders, anorexia). Families are obtained by Ward clustering of the features' standardized cross-disorder effect profiles ($k=3$ by silhouette). Every transdiagnostic feature belongs to exactly one family. Prefixes denote the source: LIWC, GoEmotions (GoEmo), or Hartmann.}
\label{tab:families}
\resizebox{\columnwidth}{!}{%
\begin{tabular}{llrrr}
\toprule
Family & Feature & dep & ed & anor \\
\midrule
\multirow{10}{*}{\shortstack[l]{F1: Reduced\\non-affective\\language\\(mean $-0.39$)}}
 & LIWC: Analytic    & $-0.57$ & $-0.36$ & $-0.76$ \\
 & LIWC: Lifestyle   & $-0.33$ & $-0.52$ & $-0.71$ \\
 & Hartmann: neutral & $-0.21$ & $-0.49$ & $-0.71$ \\
 & LIWC: Clout       & $-0.25$ & $-0.51$ & $-0.57$ \\
 & LIWC: leisure     & $-0.21$ & $-0.43$ & $-0.60$ \\
 & LIWC: money       & $-0.30$ & $-0.38$ & $-0.47$ \\
 & GoEmo: curiosity  & $-0.22$ & $-0.47$ & $-0.29$ \\
 & LIWC: QMark       & $-0.23$ & $-0.39$ & $-0.25$ \\
 & LIWC: space       & $-0.15$ & $-0.34$ & $-0.34$ \\
 & LIWC: motion      & $-0.17$ & $-0.24$ & $-0.26$ \\
\midrule
\multirow{11}{*}{\shortstack[l]{F2: Elevated\\self-focused\\affect\\(mean $+0.53$)}}
 & LIWC: i           & $+0.50$ & $+0.45$ & $+0.79$ \\
 & GoEmo: grief      & $+0.51$ & $+0.35$ & $+0.86$ \\
 & GoEmo: relief     & $+0.49$ & $+0.41$ & $+0.80$ \\
 & GoEmo: nervousness& $+0.47$ & $+0.39$ & $+0.84$ \\
 & LIWC: mental      & $+0.39$ & $+0.39$ & $+0.91$ \\
 & GoEmo: sadness    & $+0.48$ & $+0.32$ & $+0.87$ \\
 & GoEmo: caring     & $+0.51$ & $+0.39$ & $+0.69$ \\
 & LIWC: conj        & $+0.49$ & $+0.38$ & $+0.65$ \\
 & LIWC: ppron       & $+0.49$ & $+0.27$ & $+0.73$ \\
 & LIWC: pronoun     & $+0.51$ & $+0.22$ & $+0.63$ \\
 & LIWC: differ      & $+0.43$ & $+0.27$ & $+0.47$ \\
\midrule
\multirow{8}{*}{\shortstack[l]{F3: Elevated\\somatic/\\illness\\(mean $+0.44$)}}
 & LIWC: health      & $+0.32$ & $+0.51$ & $+0.89$ \\
 & Hartmann: sadness & $+0.33$ & $+0.57$ & $+0.79$ \\
 & Hartmann: fear    & $+0.23$ & $+0.54$ & $+0.67$ \\
 & LIWC: illness     & $+0.24$ & $+0.38$ & $+0.78$ \\
 & LIWC: Physical    & $+0.16$ & $+0.33$ & $+0.77$ \\
 & LIWC: feeling     & $+0.29$ & $+0.29$ & $+0.57$ \\
 & GoEmo: remorse    & $+0.25$ & $+0.20$ & $+0.55$ \\
 & LIWC: WC          & $+0.13$ & $+0.23$ & $+0.55$ \\
\bottomrule
\end{tabular}%
}
\end{table}

\begin{table}[ht]
\centering
\caption{Top-5 Bridge EI (1-step). Each ranking uses its own network's Walktrap partition.}
\label{tab:bridges}
\resizebox{0.8\columnwidth}{!}{
\begin{tabular}{lllc}
\toprule
\multirow{2}{*}{Network} & Clinical & NLP & \multirow{2}{*}{Top-5 overlap} \\
& (own partition) & (own partition) & \\
\midrule
\multirow{5}{*}{BDI}
& Concentration       & Guilt (0.69)            & \multirow{5}{*}{0/5} \\
& Sadness             & Worthlessness (0.55)    & \\
& Loss Energy         & Agitation (0.47)        & \\
& Loss Interest       & Irritability (0.46)     & \\
& Pessimism           & Loss Pleasure (0.44)    & \\
\midrule
\multirow{5}{*}{EDE-Q}
& Desire Weight Loss  & Empty Stomach (0.79)            & \multirow{5}{*}{1/5} \\
& Empty Stomach       & Social Eating Discomfort (0.70) & \\
& Shape Preoccupation & Secret Eating (0.69)            & \\
& Eating Guilt        & Shape Dissatisfaction (0.67)    & \\
& Fear Weight Gain    & Restraint (0.63)                & \\
\bottomrule
\end{tabular}%
}
\end{table}

\begin{table}[ht]
\centering
\caption{Global network metrics, uniform-Spearman EBICglasso ($\gamma = 0.5$). NLP networks are slightly denser and stronger than clinical.}
\label{tab:networks_global}
\resizebox{0.8\columnwidth}{!}{
\begin{tabular}{lcccc}
\toprule
Network & Nodes & Edges & Density & Global strength \\
\midrule
BDI clinical            & 21 & 112 & 0.533 & 9.34 \\
BDI NLP                 & 21 & 114 & 0.543 & 13.20 \\
\midrule
EDE-Q clinical          & 22 & 108 & 0.468 & 10.26 \\
EDE-Q NLP               & 22 & 124 & 0.537 & 15.47 \\
\midrule
EDE-Q subscale clinical & 4  & 6   & 1.000 & 1.79 \\
EDE-Q subscale NLP      & 4  & 5   & 0.833 & 1.86 \\
\bottomrule
\end{tabular}%
}
\end{table}

\begin{table}[ht]
\centering
\caption{Sensitivity analysis: cross-population replicating pairs at three $|\rho_\text{ED}|$ floors. All pairs additionally require FDR-significance on anorexia and same sign in both samples. The qualitative LIWC-dominant pattern is stable across the range.}
\label{tab:sensitivity}
\resizebox{\columnwidth}{!}{
\begin{tabular}{lccccc}
\toprule
Threshold & Total & \% of anor FDR-sig & GoEmo & Hartmann & LIWC \\
\midrule
$|\rho_\text{ED}| \geq 0.15$                   & 563 & 19.5\% & 128 & 34 & 398 \\
$|\rho_\text{ED}| \geq 0.20$ (pre-specified) & \textbf{242} & \textbf{8.4\%} &  40 & 21 & 180 \\
$|\rho_\text{ED}| \geq 0.25$                   &  74 &  2.6\% &  12 & 10 &  52 \\
\bottomrule
\end{tabular}%
}
\end{table}

\begin{table}[ht]
\centering
\caption{Sensitivity of the depression results to retrieval choices ($N{=}169$). Each variant changes one choice of the primary configuration. \textit{Pool $k$}: mean BGE similarity over the $k$ posts ranked highest by Stage~1 (MiniLM). \textit{Top-$k$ mean}: mean of the $k$ highest BGE similarities. \textit{Score $r$}: median per-item correlation of user scores with the primary. \textit{Edge $r$}: correlation of network edge weights with the primary NLP network and with the clinical network. Since no user exceeds 1,000 filtered posts, BGE scoring over all posts without Stage~1 is identical to the primary. All composite $\rho$ significant at $p{<}.05$ except $^\dagger$ ($p{\geq}.05$).}
\label{tab:sensitivity_retrieval}
\resizebox{\columnwidth}{!}{%
\begin{tabular}{lcccccc}
\toprule
Variant & Composite $\rho$ & FDR-sig / 21 & Score $r$ & Edges & Edge $r$ (primary) & Edge $r$ (clinical) \\
\midrule
\textbf{Primary} & \textbf{0.434} & \textbf{20} & \textbf{1.000} & \textbf{114} & \textbf{1.000} & \textbf{0.523} \\
\midrule
\multicolumn{7}{l}{\textit{Candidate pool}} \\
\quad $k=10$  & 0.222 & 6  & 0.622 & 101 & 0.776 & 0.544 \\
\quad $k=25$  & 0.193 & 2  & 0.616 & 95  & 0.809 & 0.569 \\
\quad $k=50$  & 0.205 & 1  & 0.663 & 96  & 0.850 & 0.579 \\
\quad $k=100$ & 0.246 & 13 & 0.761 & 106 & 0.915 & 0.573 \\
\quad $k=250$ & 0.377 & 19 & 0.910 & 107 & 0.969 & 0.504 \\
\midrule
\multicolumn{7}{l}{\textit{Aggregation}} \\
\quad Median        & 0.410 & 20 & 0.981 & 122 & 0.922 & 0.522 \\
\quad Maximum       & 0.161 & 0  & 0.463 & 94  & 0.700 & 0.584 \\
\quad Top-10 mean   & 0.150$^\dagger$ & 2 & 0.506 & 95 & 0.822 & 0.607 \\
\quad Top-25 mean   & 0.102$^\dagger$ & 0 & 0.461 & 94 & 0.860 & 0.592 \\
\midrule
\multicolumn{7}{l}{\textit{BDI-II query recipe}} \\
\quad Severity 1 only      & 0.432 & 20 & 0.984 & 100 & 0.756 & 0.399 \\
\quad Severity 3 only      & 0.445 & 19 & 0.984 & 117 & 0.676 & 0.429 \\
\quad Concatenated phrasings & 0.385 & 20 & 0.944 & 121 & 0.767 & 0.383 \\
\midrule
\multicolumn{7}{l}{\textit{Encoder}} \\
\quad MiniLM only & 0.362 & 16 & 0.826 & 109 & 0.551 & 0.352 \\
\bottomrule
\end{tabular}%
}
\end{table}

\begin{table}[ht]
\centering
\caption{Dimensional validation results by retrieval architecture. Composite $\rho$: Spearman correlation of the mean per-symptom symptom expression score with the questionnaire total. FDR-sig items: number of individual items reaching $q < 0.05$ after Benjamini-Hochberg correction. Sparse baselines (BM25, TF-IDF) are evaluated on the depression corpus, which has the dimensional ground truth and power for a per-item comparison. $\dagger$ = borderline ($p = 0.057$, $N = 92$, power-limited by sample size); \textsuperscript{***} $p < 0.001$; n.s. = not significant.}
\label{tab:retrieval}
\resizebox{\columnwidth}{!}{%
\begin{tabular}{lccc}
\toprule
& \multicolumn{2}{c}{Depression ($N=169$)} & ED ($N=92$) \\
\cmidrule(lr){2-3} \cmidrule(lr){4-4}
Architecture & Composite $\rho$ & FDR-sig / 21 & Composite $\rho$ \\
\midrule
BM25 (sparse)               & 0.063 (n.s.) & 0  & - \\
TF-IDF (sparse)             & 0.163 & 1  & - \\
BGE bi-encoder only         & 0.112 (n.s.) & 0  & 0.072 \\
BGE + cross-encoder         & 0.344\textsuperscript{***} & 11 & 0.186 \\
Two-stage cascade (primary) & \textbf{0.434}\textsuperscript{***} & \textbf{20} & \textbf{0.200}$\dagger$ \\
\bottomrule
\end{tabular}%
}
\end{table}

\subsection{Questionnaire Items Mapping}

The paper uses short symptom labels throughout. Table \ref{tab:bdi_mapping} maps BDI-II item numbers to the labels used in figures and prose; Table \ref{tab:edeq_mapping} does the same for EDE-Q items and subscales.

\begin{table}[ht]
\centering
\small
\setlength{\tabcolsep}{3.5pt} 
\caption{BDI-II item-to-label mapping. The label is what appears in figures, network plots, and prose throughout the paper.}
\label{tab:bdi_mapping}
\begin{tabular}{clp{4.2cm}} 
\toprule
\textbf{Item} & \textbf{Label} & \textbf{Brief description} \\
\midrule
Q1  & Sadness           & Feeling sad \\
Q2  & Pessimism         & Discouragement about the future \\
Q3  & Past Failure      & Sense of having failed \\
Q4  & Loss Pleasure     & Reduced enjoyment of activities \\
Q5  & Guilt             & Feelings of guilt \\
Q6  & Punishment        & Feeling of being punished \\
Q7  & Self Dislike      & Disappointment in self \\
Q8  & Self Blame        & Self-criticism \\
Q9  & Suicidal Ideation & Thoughts of self-harm \\
Q10 & Crying            & Frequency of crying \\
Q11 & Agitation         & Restlessness, agitation \\
Q12 & Loss Interest     & Reduced interest in others \\
Q13 & Indecisiveness    & Difficulty making decisions \\
Q14 & Worthlessness     & Feeling worthless \\
Q15 & Loss Energy       & Reduced energy \\
Q16 & Sleep Changes     & Sleep disturbance \\
Q17 & Irritability      & Increased irritability \\
Q18 & Appetite Changes  & Appetite disturbance \\
Q19 & Concentration     & Difficulty concentrating \\
Q20 & Tiredness         & Fatigue \\
Q21 & Loss Libido       & Loss of sexual interest \\
\bottomrule
\end{tabular}
\end{table}

\begin{table*}[ht]
\centering
\footnotesize 
\setlength{\tabcolsep}{2.5pt} 
\caption{EDE-Q item-to-label and subscale mapping. The 22 items used here are the symptom-content items from EDE-Q 6.0.}
\label{tab:edeq_mapping}
\begin{tabular}{clll}
\toprule
\textbf{Item} & \textbf{Label} & \textbf{Subscale} & \textbf{Brief description} \\
\midrule
Q1  & Restraint             & Restraint       & Restraint over eating \\
Q2  & Long Fasting          & Restraint       & Long periods without food \\
Q3  & Food Avoidance        & Restraint       & Avoiding particular foods \\
Q4  & Diet Rules            & Restraint       & Following dietary rules \\
Q5  & Empty Stomach         & Restraint       & Desire for empty stomach \\
Q6  & Flat Stomach          & Shape Concern   & Desire for a flat stomach \\
Q7  & Food Preoccupation    & Eating Concern  & Preoccupation with food \\
Q8  & Shape Preoccupation   & Shape Concern   & Preoccupation with shape \\
Q9  & Fear Loss Control     & Eating Concern  & Fear of losing eating control \\
Q10 & Fear Weight Gain      & Weight Concern  & Fear of weight gain \\
Q11 & Feeling Fat           & Shape Concern   & Feelings of fatness \\
Q12 & Desire Weight Loss    & Weight Concern  & Strong desire to lose weight \\
Q19 & Secret Eating         & Eating Concern  & Eating in secret \\
Q20 & Eating Guilt          & Eating Concern  & Guilt about eating \\
Q21 & Social Eating Discom. & Eating Concern  & Discomfort eating with others \\
Q22 & Weight Self Worth     & Weight Concern  & Weight updates self-eval \\
Q23 & Shape Self Worth      & Shape Concern   & Shape updates self-eval \\
Q24 & Weighing Distress     & Weight Concern  & Distress when weighing \\
Q25 & Weight Dissatisfac.   & Weight Concern  & Dissatisfaction with weight \\
Q26 & Shape Dissatisfac.    & Shape Concern   & Dissatisfaction with shape \\
Q27 & Body Vis. Discomfort  & Shape Concern   & Discomfort seeing own body \\
Q28 & Other Vis. Discomfort & Shape Concern   & Discomfort with others seeing body \\
\bottomrule
\end{tabular}
\end{table*}

\subsection{Abbreviations for Labels (Figures)}
\label{abrev_labels}

The following abbreviations were used in visuals for the BDI-II item labels: Sad = Sadness; Pess = Pessimism; PastFail = Past Failure; LossPleas = Loss Pleasure; SelfDis = Self Dislike; SelfBlm = Self Blame; SuicIde = Suicidal Ideation; Agit = Agitation; LossInt = Loss Interest; Indeci = Indecisiveness; Worthl = Worthlessness; LossEner = Loss Energy; Irrit = Irritability; Appet = Appetite Changes; Concen = Concentration; Tired = Tiredness; LossLib = Loss Libido.

The following abbreviations were used in visuals for the EDE-Q item labels: Restr = Restraint; LongFast = Long Fasting; FoodAv = Food Avoidance; DietR = Diet Rules; EmStom = Empty Stomach; FlatSt = Flat Stomach; FoodPre = Food Preoccupation; ShapePre = Shape Preoccupation; FearLC = Fear Loss Control; FearWG = Fear Weight Gain; FeelFat = Feeling Fat; DesWL = Desire Weight Loss; SecEat = Secret Eating; EatGui = Eating Guilt; SocEat = Social Eating Discomfort; WSW = Weight Self Worth; SSW = Shape Self Worth; WeighD = Weighing Distress; WDiss = Weight Dissatisfaction; SDiss = Shape Dissatisfaction; BodyVis = Body Visibility Discomfort; OthVis = Other Visibility Discomfort.

\subsection{Network Configurations}
\begin{enumerate}
    \item Network estimation: EBICglasso \citep{Friedman2008glasso} via \texttt{bootnet::estimateNetwork()} (\texttt{bootnet} v1.5.5), default tuning $\gamma=0.5$
    \item Correlations: Spearman with pairwise-complete observations, applied uniformly across all four item-level networks.
    \item Community detection: \texttt{igraph::cluster\_walktrap()} with $t=4$
    \item Bridge centrality: \texttt{networktools::bridge()} (v1.5.2), one-step Bridge Expected Influence. 
    \item Network comparison: \texttt{NetworkComparisonTest::NCT()} v2.2.2 with 1000 permutations and BH-FDR. 
    \item Centrality: \texttt{qgraph::centrality\_auto()} (\texttt{qgraph} v1.9.8).
    \item Stability: \texttt{bootnet::bootnet()} case-dropping, 1000 resamples. 
    \item Power analysis: \texttt{bootnet::netSimulator()}.
    \citep{Foygel2010}.
\end{enumerate}

\subsection{Computational Setup} All code was run on a single NVIDIA T4 GPU (16 GB) via Google Colab Pro. 

\subsection{Software versions}
Used packages and versions are as follows: Python 3.11; PyTorch 2.3.1; sentence-transformers 3.0.1; transformers 4.42.4; scikit-learn 1.5.1; pandas 2.2.2; statsmodels 0.14.2 (Benjamini-Hochberg). R 4.3.2; bootnet 1.5.5; qgraph 1.9.8; networktools 1.5.2; NetworkComparisonTest 2.2.2; igraph 2.0.3; mclust 6.1.1 (ARI).

All bootstrap and permutation procedures use a fixed seed of 42 for reproducibility. Walktrap is deterministic given the adjacency matrix.

\end{document}